\pdfoutput=1
\documentclass{article} 
\usepackage{hyperref}
\usepackage{url}

\usepackage{amsmath}
\usepackage{amssymb}
\usepackage[labelfont=bf,labelsep=period,justification=raggedright]{caption}
\usepackage[margin=35mm]{geometry}
\usepackage{graphicx} \usepackage{subfigure} \usepackage{color}
\usepackage{epstopdf} \usepackage{array} \usepackage{enumerate}
\usepackage{todonotes}
\usepackage{stmaryrd}
\usepackage{authblk}

\title{Turing Computation with\\ Recurrent Artificial Neural Networks}

\author[1]{Giovanni S.~Carmantini\thanks{giovanni.carmantini@gmail.com}}
\author[2]{Peter beim Graben}
\author[3]{Mathieu Desroches}
\author[1]{Serafim Rodrigues}
\affil[1]{\small School of Computing, Electronics and Mathematics, Plymouth University, United Kingdom}
\affil[2]{\small Bernstein Center for Computational Neuroscience Berlin, Humboldt-Universit\"at zu Berlin, Germany}
\affil[3]{\small Inria Sophia-Antipolis M{\'e}diterran{\'e}e, Valbonne, France}

\begin{document}

\maketitle

\begin{abstract}
  We improve the results by Siegelmann \& Sontag~\cite{siegelmann_computational_1995,
  siegelmann_math_1991} by providing a novel
  and parsimonious constructive mapping between Turing Machines and
  Recurrent Artificial Neural Networks, based on recent developments
  of Nonlinear Dynamical Automata. The architecture of the resulting
  R-ANNs is simple and elegant, stemming from its transparent relation
  with the underlying NDAs. These characteristics yield promise for
  developments in machine learning methods and symbolic computation
  with continuous time dynamical systems. A framework is provided to
  directly program the R-ANNs from Turing Machine descriptions, in
  absence of network training. At the same time, the network can
  potentially be trained to perform algorithmic tasks, with exciting
  possibilities in the integration of approaches akin to Google
  DeepMind's Neural Turing Machines.
\end{abstract}

\section{Introduction}
The present work provides a novel and alternative approach to the one
offered by Siegelmann and Sontag~\cite{siegelmann_computational_1995,
  siegelmann_math_1991} of mapping Turing machines to Recurrent
Artificial Neural Networks (R-ANNs). Here we employ recent theoretical
developments from symbolic dynamics enabling the mapping from Turing
Machines to two-dimensional piecewise affine-linear systems evolving
on the unit square, i.e. Nonlinear Dynamical Automata
(NDA)\cite{moore_unpredictability_1990,beim_graben_language_2004}. With
this in place, we are able to map the resulting NDA onto a R-ANN,
therefore providing an elegant constructive method to simulate a
Turing machine in real time by a first-order R-ANN.  There are two
main advantages to the proposed approach. The first one is the
parsimony and simplicity of the resulting R-ANN architecture in
respect to previous approaches. The second one is the transparent
relation between the network and its underlying piecewise
affine-linear system.  These two characteristics open the door to key
future developments when considering learning applications (see
Google DeepMind's Neural Turing Machines\cite{graves2014neural} for a relevant
example with promising future integration possibilities) -- with the exciting
possibility of a symbolic read-out of a learned algorithm from the
network weights -- and when considering extensions of the model to
continuous dynamics, which could provide a theoretical basis to query
the computational power of more complex neuronal models.

\section{Methods}
In this section we outline a mapping from Turing machines to
R-ANNs. Our construction involves two stages. In the first stage a
Generalized Shift~\cite{moore_unpredictability_1990} emulating a
Turing Machine is built, and its dynamics encoded on the unit square
via a procedure called G\"{o}delization, defining a piecewise-affine linear map on the
unit square, i.e. a NDA.
In the second stage, the resulting NDA is mapped onto a first-order R-ANN. Next, the theoretical methods employed are discussed in detail.

\subsection{Turing Machines}
A Turing Machine~\cite{turing1937computable} is a computing device
endowed with a doubly-infinite one-dimensional tape (memory support
with one symbol capacity at each memory location), a finite state
controller and a read-write head that follows the instructions encoded
by a $\delta$ transition function. At each step of the computation,
given the current state and the current symbol read by the read-write
head, the machine controller determines via $\delta$ the writing of a
symbol on the current memory location, a shift of the read-write head
to the memory location to the left ($\mathcal{L}$) or to the right
($\mathcal{R}$) of the current one, and the transition to a new state
for the next computation step. At a computation step, the content of
the tape together with the position of the read-write head and the
current controller state define a machine configuration.

More formally, a Turing Machine is a 7-tuple
$M_\text{TM} = (Q,\mathbf{N}, \mathbf{T}, q_0, \sqcup, F, \delta)$,
where $Q$ is a finite set of control states, $\mathbf{N}$ is a finite
set of tape symbols containing the blank symbol $\sqcup$,
$\mathbf{T}\subset\mathbf{N}\setminus\{\sqcup\}$ is the input
alphabet, $q_0$ is the starting state, $F\subset Q$ is a set of
`halting' states and $\delta$ is a partial transition function,
determining the dynamics of the machine. In particular, $\delta$ is
defined as follows:
\begin{equation}
  \delta:  Q\times\mathbf{N} \rightarrow
   Q\times\mathbf{N}\times\{\mathcal{L},\mathcal{R}\}.
\end{equation}
\subsection{Dotted sequences and Generalized Shifts}

A Turing machine configuration can be described by a bi-infinite
dotted sequence on some alphabet $\mathbf{A}$; it can then be defined as:
\begin{equation}
  s = \ldots d_{i_{-3}}d_{i_{-2}}d_{i_{-1}}.d_{i_{0}}d_{i_{1}}d_{i_{2}} \ldots,
\end{equation}
where $l = \ldots d_{i_{-3}}d_{i_{-2}}$ describes the part of the tape
on the left of the read-write head,
$r = d_{i_{0}}d_{i_{1}}d_{i_{2}} \ldots$ describes the part on its
right, $q=d_{i_{-1}}$ describes the current state of the machine
controller, and the dot denotes the current position of the read-write
head, i.e. the symbol to its right. The central dot splits the tape into two one-sided infinite strings $\alpha', \beta$,
where $\alpha'$ is the left part of the dotted sequence in reverse
order. The first symbol in $\alpha$ represents the current state of
the Turing Machine, whereas the first symbol in $\beta$ represents the
symbol currently under the controller's head.  The transition function
$\delta$ can be straightforwardly extended to a function $\hat{\delta}$ operating on dotted
sequences, so that
$\hat{\delta}:\mathbf{A}^\mathbb{Z}\rightarrow \mathbf{A}^\mathbb{Z}$.

A Generalized Shift acts on dotted sequences, and is defined as a pair
$M_{GS} = (\mathbf{A}^\mathbb{Z}, \Omega)$, with
$\mathbf{A}^\mathbb{Z}$ being the space of dotted sequences,
$\Omega : \mathbf{A}^\mathbb{Z} \rightarrow \mathbf{A}^\mathbb{Z}$ defined by
\begin{equation}
  \Omega(s) = \sigma^{F(s)}(s \oplus G(s))
\end{equation}
with
\begin{align}
  F&: \mathbf{A}^\mathbb{Z} \rightarrow \mathbb{Z} \\
  G&: \mathbf{A}^\mathbb{Z} \rightarrow \mathbf{A}^e
\end{align}
where $\sigma$ shifts the symbols to the left or to the right, or does not shift them
at all, as determined by the function $F(s)$. In addition, the
Generalized Shift can operate a substitution, with $G(s)$ being the
function which substitutes a substring of length $e$ in the {\it
  Domain of Effect} (DoE) of $s$ with a new substring. Both the shift
and the substitution are functions of the content of the {\it Domain
  of Dependence} (DoD), a substring of $s$ of length $\ell$.

A Turing Machine can be emulated by a Generalized Shift with
$\mbox{DoD} = \mbox{DoE} = d_{i_{-2}}d_{i_{-1}}.d_{i_0}$ and the
functions $F,G$ appropriately chosen such that
$\Omega(s)=\hat{\delta}(s)$ for all $s$ (see
\cite{moore_generalized_1991} for a detailed exposition).

\subsection{G\"{o}del codes}

G\"{o}del codes (or G\"{o}delizations) \cite{Goedel31} map strings 
to numbers and, in particular, allow the mapping of the space
of one-sided infinite sequences to the real interval $[0, 1]$.
Let $\mathbf{A}^\mathbb{N}$ be the space of one-sided infinite
sequences over an alphabet $\mathbf{A}$,
$s$ be an element of $\mathbf{A}^{\mathbb{N}}$, $r_k$
the $k$-th symbol in $s$, $\gamma: \mathbf{A}\rightarrow\mathbb{N}$ a
one-to-one function associating each symbol in the alphabet
$\mathbf{A}$ to a natural number, and $g$ the number of symbols in
$\mathbf{A}$.  Then a G\"{o}delization is a mapping $\psi$ from
$\mathbf{A}^\mathbb{N}$ to $[0,1]\subset \mathbb{R}$ defined as:
\begin{equation}
  \psi(s) := \sum\limits_{k=1}^{\infty} \gamma(r_k)g^{-k}.
  \label{eq:godel_encoding}
\end{equation}
Conveniently, G\"{o}delization can be employed on a Turing machine
configuration, represented as a dotted sequence
$\alpha.\beta \in \mathbf{A}^\mathbb{Z}$. 
The G\"{o}del encoding $\psi_{x}$ and $\psi_{y}$ of $\alpha'$ and
$\beta$ define a representation of $s$
$(\psi_{x}(\alpha'), \psi_{y}(\beta))$ known as symbol plane or
symbologram representation, which is contained in the unit square
${[0,1]}^2 \subset \mathbb{R}^2$. The choice of encoding
$\psi_{x}$ and $\psi_{y}$ to use on the machine configurations is arbitrary. Therefore, to enable the construction of
parsimonious Nonlinear Dynamical Automata our encoding will assume
that $\beta$ always contains tape symbols only, and that the first
symbol of $\alpha'$ is always a state symbol, the rest being tape
symbols only. Based on these assumptions, the particular
encoding is defined as:
\begin{equation}
  \begin{aligned}
    \psi_x(\alpha') &= \gamma_q(a_1)n_q^{-1} +
    \sum\limits_{k=1}^{\infty} \gamma_s(a_{k+1})n_s^{-k}n_q^{-1},\\
    \psi_y(\beta) &= \sum\limits_{k=1}^{\infty} \gamma_s(b_k)n_s^{-k},
  \end{aligned}
  \label{eq:encodings}
\end{equation}
with $n_q = \lvert Q \rvert$, i.e. the number of states in the Turing
Machine, $n_s = \lvert \mathbf{N} \rvert$, i.e. the number of tape
symbols in the Turing Machine, $\gamma_q$ and $\gamma_s$ enumerating
$ Q$ and $\mathbf{N}$ respectively, and with $a_k$ and $b_k$ being the
$k$-th symbol in $\alpha'$ and $\beta$ respectively.

\subsubsection{Encoded Generalized Shift and affine-linear transformations}\label{codes}

The substitution and shift operated by a Generalized Shift on a dotted
sequence $s = \alpha . \beta$ can be represented as an affine-linear
transformation on $(\psi_x(\alpha'), \psi_y(\beta))$, i.e.~the
symbologram representation of $s$. In particular, a substitution and
shift on a dotted sequence can be broken down into substitutions and
shifts on its one-sided components. In the following, we will show how
substitutions and shifts on a one-sided infinite sequence can be
represented as affine-linear transformations on its
G\"{o}delization. These results will be useful in showing how the
symbologram representation of a Generalized Shift leads to a
piecewise affine-linear map on a rectangular partition of the unit square.\\
Let $s = d_1 d_2 d_3 \ldots$ be a one-side infinite sequence on some
alphabet $\mathbf{A}$. Substituting the $n$-th symbol in $s$ with
$\hat{d}_n$ yields
$\hat{s} = d_1 \ldots d_{n-1} \hat{d}_n d_{n+1}\ldots$, so that
\begin{align*}
  \psi(s) & =  \gamma(d_1)g^{-1} + \ldots
                  \gamma(d_{n-1})g^{-(n-1)} +
                  \gamma(d_n)g^{-n} + \gamma(d_{n+1})g^{n+1} + \ldots,\\
  \psi(\hat{s}) & = \gamma(d_1)g^{-1} + \ldots
                        \gamma(d_{n-1})g^{-(n-1)} +
                        \gamma(\hat{d}_n)g^{-n} + \gamma(d_{n+1})g^{n+1} + \ldots,\\
  & = \psi(s) - \gamma(d_n)g^{-n} + \gamma(\hat{d}_n)g^{-n}.
\end{align*}
As the previous example illustrates, G\"{o}delizing a sequence resulting from a
symbol substitution is equivalent to applying an affine-linear transformation
on the original G\"{o}delized sequence. In particular, the parameters
of the affine-linear transformation only depend on the position and
identities of the symbols involved in the substitution.
Shifting $s$ to the left by removing its first symbol or shifting it
to the right by adding a new one yields respectively
$s_l = d_2 d_3 d_4 \ldots$ and $s_r = b\; d_1 d_2 d_3 d_4 \ldots$,
where $b$ is the newly added symbol. In this case
\begin{align*}
  \psi(s_l) & = \gamma(d_2)g^{-1} + \gamma(d_3)g^{-2} +
              \gamma(d_4)g^{-3} + \ldots\\
            & = g \psi(s) - \gamma(d_1),\\
  \intertext{and}
  \psi(s_r) & = \gamma(b)g^{-1} + \gamma(d_1)g^{-2} +
              \gamma(d_2)g^{-3} + \gamma(d_3)g^{-4} + \ldots\\
            & = g^{-1} \psi(s) + \gamma(b)g^{-1}.
\end{align*}

Again, the resulting G\"{o}delized shifted sequence can be obtained by
applying an affine-linear transformation to the original G\"{o}delized
sequence.

\subsection{Nonlinear Dynamical Automata}

A Nonlinear Dynamical Automaton (NDA) is a triple
$M_{NDA} = (X, P, \Phi)$, with $P$ being a rectangular partition of the
unit square, that is
\begin{equation}
  P = \{D^{i,j} \subset X |~ 1 \le i \le m,\; 1 \le j \le n,\; \space m,n \in \mathbb{N}\},
  \label{Eq:partition}
\end{equation}
so that each cell $D^{i,j}$ is defined as the cartesian product $I_i \times J_j$, with
$I_i, J_j \subset [0,1]$ being real intervals for each bi-index
$(i,j)$, $D^{i,j} \cap D^{k,l} = \emptyset$ if
$(i,j) \neq (k,l)$, and $\bigcup_{i,j}D^{i,j}=X$.\\
The couple $(X, \Phi)$ is a time-discrete dynamical system with phase
space $X={[0, 1]}^2 \subset \mathbb{R}^2$ (i.e.~the unit square) and
with flow $\Phi: X \rightarrow X$, a piecewise affine-linear map such
that $\Phi_{|D^{i,j}}:=\Phi^{i,j}$. Specifically, $\Phi^{i,j}$ takes the following
form:
\begin{equation}
  \Phi^{i,j}(\mathbf{x}) = \left(\begin{array}{c} a^{i,j}_x\\ a^{i,j}_y \end{array}\right) +
  \left(\begin{array}{cc} \lambda^{i,j}_x & 0 \\ 0 & \lambda^{i,j}_y \end{array}\right)
  \left(\begin{array}{c} x\\ y \end{array}\right).
  \label{Eq:NDA_dynamics}
\end{equation}
The piecewise affine-linear map $\Phi$ also requires a switching rule
$\Theta(x,y) \in \llbracket 1,m \rrbracket \times \llbracket 1,n
\rrbracket$
to select the appropriate branch, and thus the appropriate dynamics,
as a function of the current state. That is,
$\Phi(x,y) = \Phi^{i,j}(x, y) \iff \Theta(x,y) = (i,j)$.

Each cell $D^{i,j}$ of the partition $P$ of the unit square can be
seen as comprising all the G\"{o}delized dotted sequences that
contain the same symbols in the Domain of Dependence. That is, for a
Generalized Shift simulating a Turing Machine, the first two symbols
in $\alpha'$ and the first symbol in $\beta$.\\
The unit square is thus partitioned in a number of $I$ intervals equal
to $m = n_q n_s$, and one of $J$ intervals equal to $ n = n_s$, with
$n_q$ being the number of states in $Q$ and $n_s$ the number of
symbols in $\mathbf{N}$, for a total of $n_q n_s^2$ cells. As each cell
corresponds to a different Domain of Dependence of the underlying
Generalized Shift in symbolic space, it is associated with a different
affine-linear transformation representing the action of a substitution
and shift in vector space. The transformation parameters
$(a^{i,j}_x, a^{i,j}_y)$ and $(\lambda^{i,j}_x, \lambda^{i,j}_y)$ can
be derived using the methods outlined in~\autoref{codes}.\\
Thus, a Turing Machine can be represented as a Nonlinear
Dynamical Automaton by means of its G\"{o}delized Generalized Shift
representation.

\section{NDAs to R-ANNs}
The aim of the second stage of our methodology is to map the orbits of
the NDA (i.e. $\Phi^{i,j}(x,y)$) to orbits of the R-ANN, which we will
denote by $\zeta^{i,j}(x,y)$.

Let $\rho(\cdot)$ denote the proposed map. Its role is to encode the
affine-linear dynamics at each $\Phi^{i,j}$ branch in the architecture
and weights of the network, and emulate the overall dynamics $\Phi$ by
suitably activating certain neural units within the R-ANN given the
switching rule $\Theta$. Therefore, we generically define the proposed
map as follows:
\begin{equation}
  \zeta = \rho(\mathcal{I}, \mathcal{A}, \Phi, \Theta),
\end{equation}
where $\mathcal{I}$ is the identity matrix mapping
(identically) the initial conditions of the NDA to the R-ANN and
$\mathcal{A}$ is the adjacency matrix specifying the network
architecture and weights, which will be explained in subsequent
sections. In addition, $\rho$ defines different neural dynamics for
each type of the neural units, that is,
$\zeta=(\zeta_1,\zeta_2,\zeta_3)$ corresponding to MCL,
BSL and LTL, respectively (see below for the definitions of these acronyms). The details of the R-ANN architecture and its dynamics
are subsequently discussed.

\subsection{Network architecture and neural dynamics}

\begin{figure}

  \centering
    {\includegraphics[width=4in]{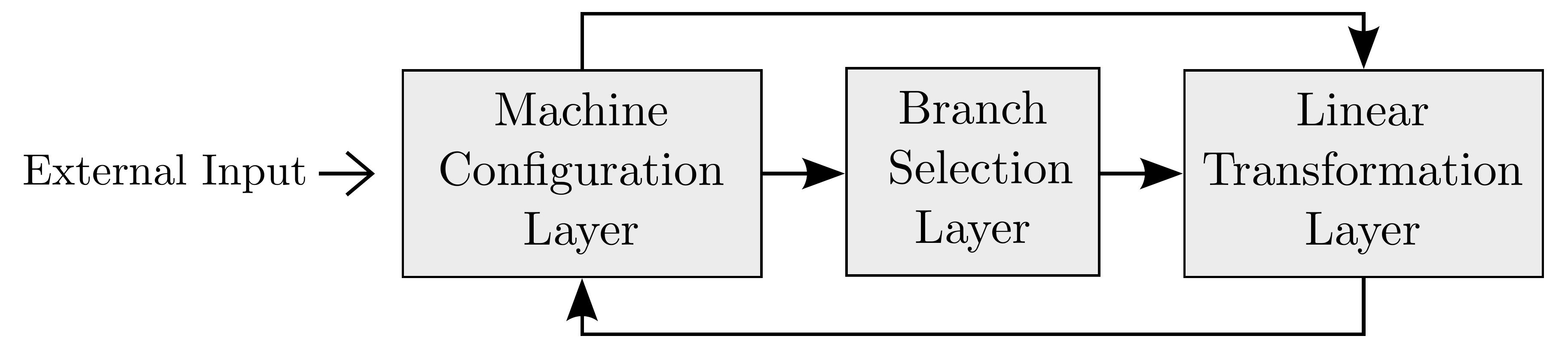}}

    \caption{Connectivity between neural layers within the
      network.}
    \label{fig:network_architecture}
\end{figure}

The proposed map,  $\rho$, attempts to mirror the affine-linear dynamics
(given by~\autoref{Eq:NDA_dynamics}) of an NDA on the partitioned
unit square (see~\autoref{Eq:partition}) by endowing the R-ANN with a
structure capturing the characteristic features of a piecewise-affine
linear system, i.e.~a state, a switching rule and a set of
transformations.\\
To achieve this, we propose a network architecture
with three layers, namely a Machine Configuration Layer (MCL) encoding
the state, a Branch Selection Layer (BSL) implementing the switching
rule and a Linear Transformation Layer (LTL), as depicted
in~\autoref{fig:network_architecture}.\\
The neural units within the various layers make use of either the
Heaviside ($H$) or the Ramp ($R$) activation functions defined as
follows:

\noindent
\begin{minipage}{.5\linewidth}
  \begin{equation}
    H(x)=
    \begin{cases}
      0 &\mbox{if } x<0\\
      1 &\mbox{if } x\ge0\\
    \end{cases}
    \label{Eq:Heaviside}
  \end{equation}
\end{minipage}
\begin{minipage}{.5\linewidth}
  \begin{equation}
    R(x) =
    \begin{cases}
      0 &\mbox{if\;} x < 0\\
      x &\mbox{if\;} x \ge 0\\
    \end{cases}.
    \label{Eq:Ramp}
  \end{equation}
\end{minipage}

Since $\Phi$ is a two-dimensional map, this suggests only two neural
units ($c_x, c_y$) in the MCL layer encoding its state at every step.
A set of BSL units functionally acts as a switching system that
determines in which cell $D^{i,j}$ the current Turing machine
configuration belongs to and then triggers the specific LTL unit
emulating the application of an affine-linear transformation
$\Phi^{i,j}$ on the current state of the system. The result of the
transformation is then fed back to the MCL for the next iteration.
On the symbolic level, one iteration of the emulated NDA corresponds
to a tape and state update of the underlying Turing machine, which can
be read out by decoding the activation of the MCL neurons.

\subsubsection{Machine Configuration Layer}
The role of the MCL is to store the current G\"{o}delized
configuration of the simulated Turing Machine at each computation
step, and to synaptically transmit it to the BSL and LTL layers. The layer
comprises two neural units ($c_x$ and $c_y$), as needed to store
the G\"{o}delized dotted sequence representing a Turing Machine
configuration (see~\autoref{eq:encodings}).

\begin{figure}
  \centering \subfigure[Branch Selection
  Layer\label{fig:Branch_Selection_Layer}]{
    \includegraphics[width=.4\linewidth]{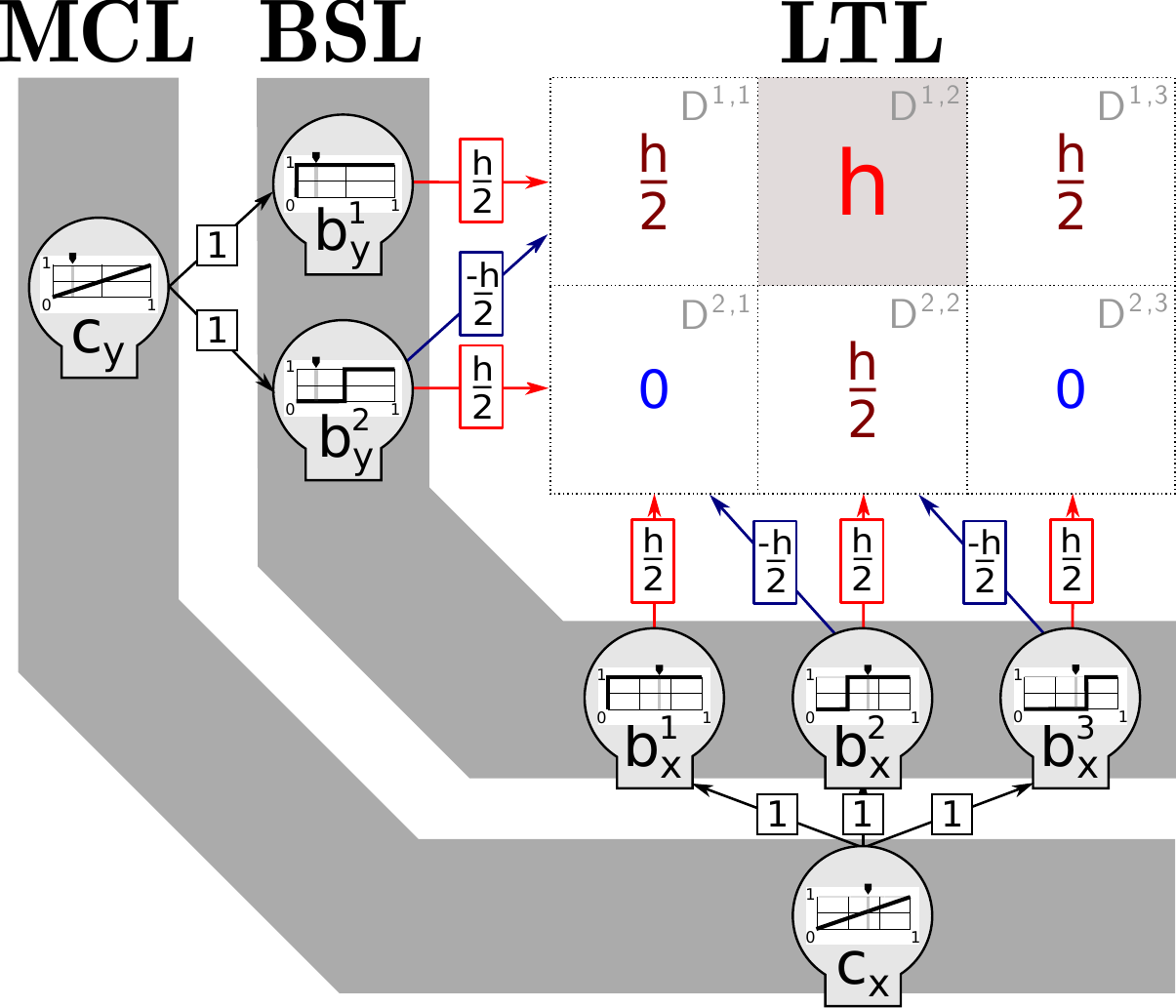}}
  \hspace*{.5em}
  \subfigure[Complete branch connection layout\label{fig:complete_branch}]{
    \includegraphics[width=.4\linewidth]{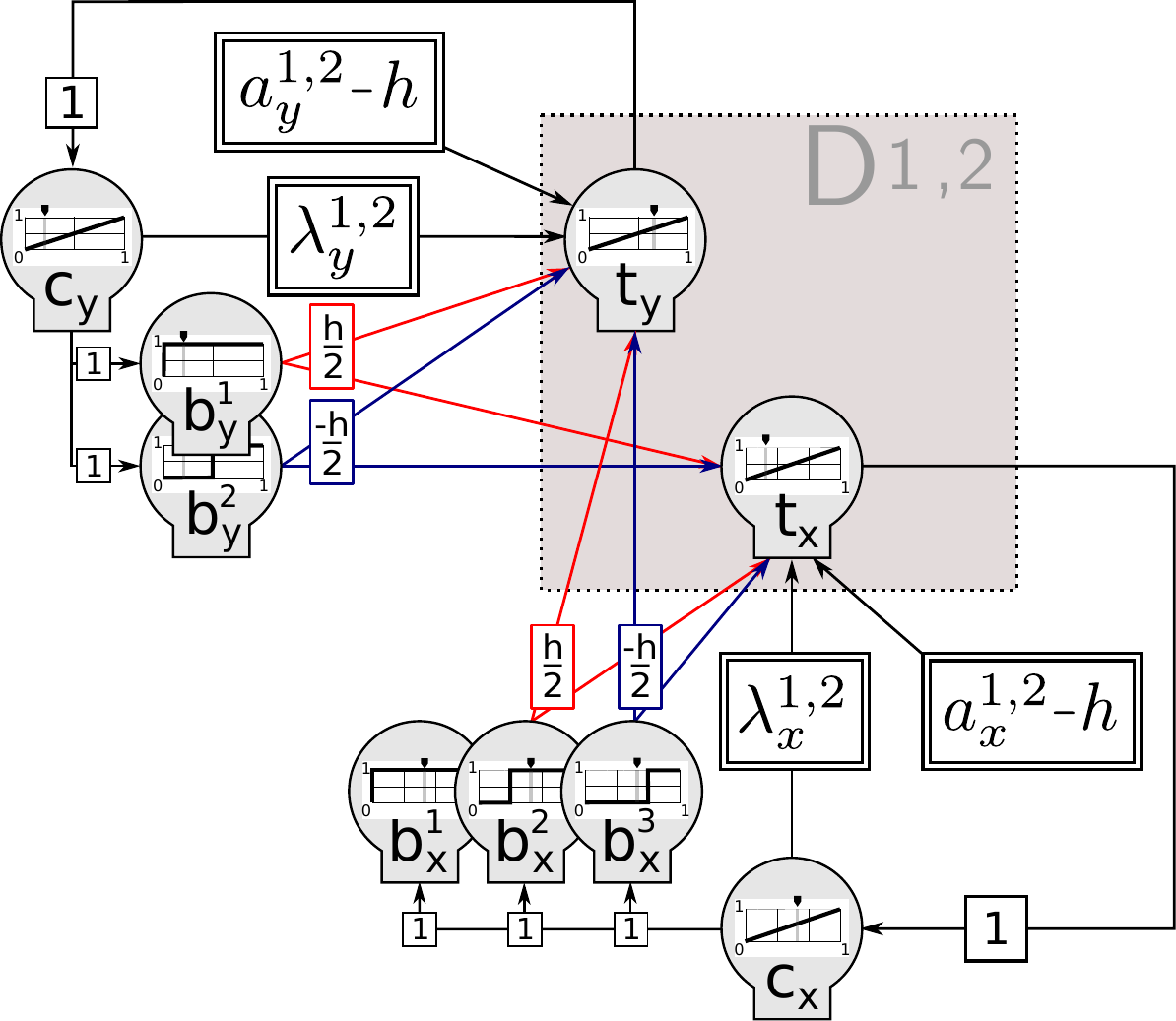}}
  \caption{Detailed feedforward connectivity and weights for a neural
    network simulating a NDA with only 6 branches.}
  \label{fig:full_BSL}
\end{figure}

The R-ANNs is thus initialized by activating this layer, given the NDA
initial conditions $(\psi_x(\alpha'), \psi_x(\beta))$ which are
identically transformed via $\mathcal{I}$ by the map $\rho(\cdot)$ as
follows:
\begin{equation}
  (c_x, c_y) =  (\psi_x(\alpha'), \psi_x(\beta)) \equiv \zeta_1 = \rho(\mathcal{I}, \cdot , \cdot, \cdot)_{| (\psi_x(\alpha'), \psi_x(\beta))}
\end{equation}
At each iteration, the units in this layer receive input from the LTL
units, and are activated via the ramp activation function
(\autoref{Eq:Ramp}); in other words
$\zeta_1 \equiv (c_x, c_y )= (R(\sum_i t^{i}_x), R(\sum_j t^{j}_y)$).
Finally, the MCL synaptically projects onto the BSL and LTL (refer to
\autoref{fig:full_BSL} for details of the connectivity).

\subsubsection{Branch Selection Layer}
The BSL embodies the switching rule $\Theta(x,y)$ and
coordinates the dynamic switching between LTL units.
In particular, if at the current step the MCL activation is
$(c_x, c_y) \in D^{i,j} = I_i \times J_j$, with
$I_i = [\xi_i, \xi_{i + 1})$ being the $i$-th interval on the $x$-axis
and $J_j = [\eta_j, \eta_{j + 1})$ being the $j$-th interval on the
$y$-axis, the BSL units activate only the $(t^{i,j}_x, t^{i,j}_y)$
units in the LTL. In this way, only one couple of LTL units is active at
each step. The switching rule is mapped by $\rho(\cdot)$ as follows:
\begin{eqnarray}
  \zeta_{2}(x,y) = \rho(\cdot, \cdot, \cdot, \Theta(x,y)=(i,j)).
\end{eqnarray}
The BSL is composed of two groups of Heaviside
(\autoref{Eq:Heaviside}) units, implementing respectively the $x$ and
the $y$ component of the switching rule of the underlying piecewise
affine-linear system, namely: i) the $b_x$ group receives input with weight
$1$ from the $c_x$ unit of the MCL layer, and comprises $n_q n_s$
units (i.e.  $b^{i}_x, 1 \leq i \leq n_q n_s$); ii) the $b_y$ group
receives input with weight $1$ from $c_y$ and comprises $n_s$ units
(i.e.  $b^{j}_y, 1 \leq i \leq n_s$). The activation of the two groups
of units is defined as:
\begin{equation}
  \begin{aligned}
    b_x^i & = H(c_x - \xi^i) \quad && \text{with} & \quad \xi^i & = \min(I_i),\\
    b_y^j & = H(c_y - \eta^j) \quad && \text{with} & \quad \eta^j & = \min(J_i).
  \end{aligned}
\end{equation}
Each $b_x^{i}$ and $b_y^{j}$ BSL unit has an activation threshold,
defined as the left boundary of the $I_i$ and $J_j$
intervals, respectively, and implemented as input from an always-active bias unit
(with weight $-\xi^i$ for the $b_x^i$ unit and $-\eta^j$ for
$b_y^j$). Therefore, an activation of $(c_x, c_y)$ in the MCL
corresponding to a point on the unit square belonging to cell
$D^{i,j}$, would trigger active all units $b_x^k$ with $k \le i$. The
same would occur for all neural units $b_y^k$ with
$k \le j$.\footnote{Note that the action of the BSL could be
  equivalently implemented by interval indicator functions represented
  as linear combinations of Heaviside functions.}

Each $b_x^i$ unit establishes synaptic excitatory connections (with
weight $\frac{h}{2}$) to all LTL units corresponding to cells
$D^{k,i}$ (i.e. $(t_x^{k,i}, t_y^{k,i})$) and inhibitory connections
(with weight $\frac{-h}{2}$) to all LTL units corresponding to cells
$D^{k,i-1}$ (i.e.  $(t_x^{k,i-1}, t_y^{k,i-1})$), with
$k=1, \ldots , n_s$; for a graphical representation
see~\autoref{fig:full_BSL}. Similarly, each $b_y^j$ unit establishes
synaptic excitatory connections to all LTL units corresponding to
cells $D^{j,k}$ and inhibitory connections to all LTL units
corresponding to cells $D^{j-1,k}$, with $k=1, \ldots , n_qn_s$.
Together, the $b^{i}_x$ and $b^{j}_y$ units completely counterbalance
through their synaptic excitatory connections the natural inhibition
(of bias $h$, which value and definition will be discussed in the
following section) of the LTL units corresponding to cell $D^{i,j}$
(i.e. $(t_x^{i,j}, t_y^{i,j})$).

In other words each couple of LTL units $(t_x^{i,i}, t_y^{i,j})$
receives an input of $B_x^i + B_y^j$, defined as follows:
\begin{equation}
  \begin{aligned}
    B_x^i &&=&& b_x^i \frac{h}{2} + b_x^{i+1} \frac{-h}{2},\\
    B_y^j &&=&& b_y^j \frac{h}{2} + b_y^{j+1} \frac{-h}{2},
  \end{aligned}
  \label{eq:Bi+Bj_definition}
\end{equation}
where the input sum
\begin{equation}
  B_x^i + B_y^j =
  \begin{cases}
    h &\mbox{ if } (c_x, c_y) \in D_{i,j}\\
    \frac{h}{2} & \mbox{ if } c_x \in I_i, c_y \not\in J_j \quad \mbox{ or }\quad c_x \not\in I_i, c_y \in J_j\\
    0 & \mbox{ if } (c_x, c_y) \not\in D_{i,j}
  \end{cases}
  \label{eq:Bi+Bj_value}
\end{equation}
only triggers the relevant LTL unit if it reaches the value $h$. That
is, if $(c_x, c_y) \in D_{i,j}$ then $B_x^i + B_y^j = h$, and the pair
$(t^{i,i}_x, t^{i,j}_y)$ is selected by the BSL units. Otherwise
$(t^{i,i}_x, t^{i,j}_y)$ stays inactive as $B_x^i + B_y^j$ is either
equal to $\frac{h}{2}$ or $0$, which is not enough to win the LTL pair natural
inhibition. An example of this mechanism is shown
in~\autoref{fig:full_BSL} , where the LTL units in cell $D^{1,2}$ are
activated via mediation of $b_x= \{b^{1}_x, b^{2}_x, b^{3}_x\}$ and
$b_y = \{b^{1}_y, b^{2}_y\}$.  Here, both $b^{3}_x$ and $b^{2}_y$ are
not excited since $c_x$ and $c_y$, respectively, are not activated
enough to drive them towards their threshold.  However, $b_x^{2}$
excites (with weights $\frac{h}{2}$) the LTL units in cell $D^{2,2}$
and $D^{1,2}$ and inhibits (with weights $\frac{-h}{2}$) the LTL units
in cell $D^{2,1}$ and $D^{1,1}$. Equally, $b_y^{2}$ excites (with
weights $\frac{h}{2}$) the LTL units in cell $D^{2,1}$, $D^{2,2}$ and
$D^{2,3}$ and inhibits (with weights $\frac{-h}{2}$) the LTL units in
cells $D^{1,1}$, $D^{1,2}$ and $D^{1,3}$.  The $b_x^{1}$ and $b_y^{1}$
units excite cells $\{D^{2,1}, D^{1,1}\}$ and
$\{D^{1,1}, D^{1,2}, D^{1,3} \}$, respectively, but these do not inhibit any cells
(due to boundary conditions).

\subsubsection{Linear Transformation Layer}
The LTL layer can be functionally divided in sets of two units, where
each couple applies two decoupled affine-linear transformations
corresponding to one of the branches of the simulated NDA. On the
symbolic level, this endows the LTL with the ability to generate an
updated machine configuration from the previous one. In the LTL, a
branch $(i,j)$ of a NDA,
$\Phi^{i,j}(x,y) =( \lambda^{i,j}_x x + a^{i,j}_x, \lambda^{i,j}_y y+
a^{i,j}_y)$,
is simulated by the LTL units $(t^{i,j}_x,t^{i,j}_y)$. Mathematically,
this induces the following mapping:
\begin{eqnarray}
  (t^{i,j}_x, t^{i,j}_y) &=& \zeta_{3}^{i,j}(x,y) = \rho(\cdot, \cdot, \Phi^{i,j}(x,y), \cdot).
\end{eqnarray}
The affine-linear transformation is implemented synaptically, and it is
only triggered when the BSL units provide enough excitation to enable
$(t^{i,j}_x, t^{i,j}_y)$ to cross their threshold value and execute
the operation. The read-out of this process corresponds to:
\begin{equation}
  \begin{array}{lcl}
    t^{i,j}_x &=& R(\lambda^{i,j}_x c_x + a^{i,j}_x - h + B_x^i + B_y^j),\\
    t^{i,j}_y &=& R(\lambda^{i,j}_y c_y + a^{i,j}_y - h + B_x^i + B_y^j).
  \end{array}
  \label{eq:LTL}
\end{equation}
A strong inhibition bias $h$ (implemented as a synaptic projection
from a bias unit) plays a key role in rendering the LTL units inactive
in absence of sufficient excitation. The bias value is defined as
follows
\begin{equation}
  - \frac{h}{2} \leq -\max_{i,j,k}(a^{i,j}_k + \lambda^{i,j}_k) \quad \text{with} \quad k=\{x,y\}.
  \label{eq:inhibition}
\end{equation}\\
Hence, each of the BSL inputs $B^{i}_x$ and $B^{i}_y$ contributes
respectively to half of the necessary excitation ($\frac{h}{2}$)
needed to counterbalance the LTL's natural inhibition (refer to
\autoref{eq:Bi+Bj_definition} and~\autoref{eq:Bi+Bj_value}).

The LTL units receive input from the two CSL units $(c_x, c_y)$, with
synaptic weights of $(\lambda^{i,j}_x, \lambda^{i,j}_y)$, and they are also
endowed with an intrinsic constant LTL neural dynamics
$(a^{i,j}_x, a^{i,j}_y)$. If the input from the BSL layer is enough
for these neurons to cross the threshold mediated by the Ramp
activation function, the desired affine-linear transformation is
applied. The read-out is an updated encoded Turing machine
configuration, which is then synaptically fed back to the CSL units
$(c_x, c_y)$, ready for the next iteration (or next Turing machine
computation step on the symbolic level).

\subsubsection{NDA-simulating first order R-ANN}

The NDA simulation (and thus Turing machine simulation) by the R-ANN
is achieved by a combination of synaptic and neural computation among the
three neural types (MCL, BSL, and LTL) and with a total of
\begin{equation}
  n_\text{units} = \underbrace{\vphantom{n_q}2}_\text{MCL} + \underbrace{n_s + n_s n_q}_\text{BSL} +
  \underbrace{2 n^2_s n_q}_\text{LTL} + \underbrace{\vphantom{n_q}1}_\text{bias unit}
  \label{eq:n_units}
\end{equation}
neural units, where $n_q$ and $n_s$ are the number of states and the
number of symbols in the Turing Machine to be simulated,
respectively. These units are connected as specified by an adjacency
matrix $\mathcal{A}$ of size $n_\text{units} \times n_\text{units}$,
following the connectivity pattern described
in~\autoref{fig:network_architecture} and with synaptic weights as
entries from the set
$$\{0, 1, \frac{h}{2}, \frac{-h}{2}\} \cup \{a^{i,j}_k - h \; \vert \;
i=1,\ldots,n_qn_s \enskip,\enskip j=1,\ldots,n_s \enskip, \enskip
k=x,y\},$$ the second component being the set of biases.

An important modelling issue to consider is that of the halting
conditions for the ANN, i.e. when to consider the computation
completed. In the original formulation of the Generalized Shift, there is
no explicit definition of halting condition. As our ANN model is based
on this formulation, a deliberate choice has to be made in its
implementation. Two choices seem to be the most reasonable. The first
one involves the presence of an external controller halting the
computation when some conditions are met, i.e. an \emph{homunculus}
\cite{beim_graben_language_2004}. The second one is the implementation
of a fixed point condition, intrinsic to the dynamical system,
representing a TM halting state as an Identity branch on the NDA. In this
way a halting configuration will result in a fixed point on the NDA,
and thus on the R-ANN. In other words, the network's computation is
considered completed if and only if
\begin{equation}
  \zeta_1(x',y') = (x', y').
  \label{eq:halting_condition}
\end{equation}
In the present study we decided to use a fixed point halting
condition, but the use of a \emph{homunculus} would likely be more
appropriate in other contexts such as interactive computation
\cite{beim_graben_quantum_2008, beim_graben_towards_2008, Wegner98}
or cognitive modelling, where different kinds of fixed points are
required in order to describe sequential decision problems
\cite{RabinovichEA08}, such as linguistic garden paths
\cite{beim_graben_language_2004, beim_graben_towards_2008}.

The implementation of the R-ANN defined like so simulates a NDA in
real-time and, thus, it simulates a Turing Machine in real time. More
formally, it can be shown that under the map $\rho(\cdot)$ the
commutativity property $\zeta \circ \rho = \rho \circ \Phi$ is
satisfied, which extends the previously demonstrated commutativity
property between Turing machines and
NDAs~\cite{beim_graben_quantum_2008,beim_graben_inverse_2009,beim_graben_universal_2014}.

\section{Discussion}
In this study we described a novel approach to the mapping of Turing
Machines to first-order R-ANNs. Interestingly, R-ANNs can be
constructed to simulate any piecewise affine-linear system on a
rectangular partition of the $n$-dimensional hypercube by extending
the methods discussed

The proposed mapping allows the construction, given any Turing
Machine, of a R-ANN simulating it in real time. As an example of the
parsimony we claim, a Universal Turing Machine can be simulated with a
fraction of the units than previous approaches allowed for: the
proposed mapping solution derives a R-ANN that can simulate Minsky's
7-states 4-symbols UTM \cite{minsky1962size} in real-time with 259
units (as per \autoref{eq:n_units}), approximately 1/3 of the 886
units needed in the solution proposed by Siegelmann and
Sontag~\cite{siegelmann_computational_1995}, and with a much simpler
architecture.

In future work we plan to overcome some of the issues posed by the
mapping and parts of its underlying theory, especially in relation to
learning applications. Key issues to overcome are the missing
end-to-end differentiability, and the need for a de-coupling of states
and data in the encoding. A future development would see the
integration of methods of data access and manipulation akin to that in
Google DeepMind's Neural Turing Machines \cite{graves2014neural}.  A
parallel direction of future work would see the mapping of Turing
machines to continuous-time dynamical systems (an example with
polynomial systems is provided in~\cite{graca_etal_08}). In
particular, heteroclinic dynamics~ \cite{RabinovichEA08, beim_graben_inverse_2009,
  Tsuda01, krupa1997robust} -- with machine
configurations seen as metastable states of a dynamical system -- and
slow-fast dynamics~\cite{desroches2013inflection,
  desroches2015sirev} are promising new directions of research.

\subsection*{References}
\begingroup
\renewcommand{\section}[2]{} 
\bibliography{bibliography}
\endgroup

\end{document}